\documentclass[conference]{IEEEtran}
\IEEEoverridecommandlockouts
\usepackage{cite}
\usepackage{amsmath,amssymb,amsfonts}
\usepackage{algorithm}
\usepackage{algpseudocode}
\usepackage{graphicx}
\usepackage{textcomp}
\usepackage{xcolor}
\usepackage{geometry}
\usepackage{booktabs,siunitx}
\usepackage{amssymb}
\usepackage{colortbl}
\usepackage{booktabs}
\usepackage{url}

\def\BibTeX{{\rm B\kern-.05em{\sc i\kern-.025em b}\kern-.08em
    T\kern-.1667em\lower.7ex\hbox{E}\kern-.125emX}}
\begin{document}

\title{Landmark Stereo Dataset for Landmark Recognition and Moving Node Localization in a Non-GPS Battlefield Environment\\

}
\author{\IEEEauthorblockN{Ganesh Sapkota}
\IEEEauthorblockA{\textit{Department of Computer Science} \\
\textit{Missouri University of Science and Technology}\\
Rolla,MO,USA \\
gs37r@mst.edu}
\and
\IEEEauthorblockN{Sanjay Madria}
\IEEEauthorblockA{\textit{Department of Computer Science} \\
\textit{Missouri University of Science and Technology}\\
Rolla,MO,USA\\
madrias@mst.edu}
}

\maketitle

\begin{abstract}
In this paper, we have proposed a new strategy of using the landmark anchor node instead of a radio-based anchor node to obtain the virtual coordinates (landmarkID, DISTANCE) of moving troops or defense forces that will help in tracking and maneuvering the troops along a safe path within a GPS-denied battlefield environment. The proposed strategy implements landmark recognition using the Yolov5 model and landmark distance estimation using an efficient Stereo Matching Algorithm. We consider that a moving node carrying a low-power mobile device facilitated with a calibrated stereo vision camera that can capture stereo images of a scene containing landmarks within the battlefield region whose locations are stored in an offline server residing within the device itself. We created a custom landmark image dataset called MSTLandmarkv1 with 34 landmark classes and another landmark stereo dataset of those 34 landmark instances called MSTLandmarkStereov1. We trained the YOLOv5 model with MSTLandmarkv1 dataset and achieved 0.95 mAP @ 0.5 IoU and 0.767 mAP @ [0.5: 0.95] IoU. We calculated the distance from a node to the landmark utilizing the bounding box coordinates and the depth map generated by the improved SGM algorithm using MSTLandmarkStereov1. The tuple of landmark IDs obtained from the detection result and the distances calculated by the SGM algorithm is stored as the virtual coordinate of a node. In future work, we will use these virtual coordinates to obtain the location of a node using an efficient trilateration algorithm and optimize the node position using the appropriate optimization method.

\end{abstract}


\section{Introduction}
A particularly important application of establishing WSNs on the battlefield is sensing, and localization of moving objects (e.g. defense forces, troops, war vehicles, opponents troops) along any trajectory\cite{wsn_battlefield_application,battlefield_surv_using_WSN}. Applications of localization could be on tracking the node movement, predicting its future trajectory and sometimes guiding the node along the safe trajectory to increase survivability during harsh situations. WSN in the battlefield includes but is not limited to several active and passive sensors that are deployed in concealed observation posts or carried by the reconnaissance units. Such sensors are often used to detect and examine Chemical, Biological, Radiological, and Nuclear (CBRN) threats. Although the network of sensors communicating with the mobile reconnaissance unit could track the activities of the adversary and report it to the intelligence center promptly and accurately, those signals and radio footprints could be eavesdropped on and detected by the opponents. So, Battlefields have become overwhelmingly contentious for precious wireless resources and satellite communications, thus making the ability to establish wireless sensor networks (WSNs) difficult as well as dangerous for necessary components\cite{anchor_node_localization}. Also, the commonly accepted assumption for mission planners is that Global Positioning System (GPS) connectivity may be either intermittent or entirely unavailable in some Named Areas of Interest (NAIs) due to unusual terrain. The direct exchange of the Global Positioning System (GPS) information between sensors, forces, troop leaders, or reconnaissance squadron units is riskier even though they are transmitted through encrypted channels. They may still expose information about NAIs, which may increase the risk of opponent attacks or direct firing within the NAIs and may prevent the defense forces from occupying those areas within the battlefield.\\
Given these restrictions and limitations of exchanging GPS included radio footprints between sensors, defense forces, and reconnaissance squadrons, much research has been conducted on alternate methods of establishing and maintaining non-GPS based WSN Framework in battlefields including range-free approach such as Distance-Vector Hop (DV Hop) algorithms \cite{range_free_localization} that utilizes physically deployed anchor nodes, regular sensor nodes, and virtual coordinate based on the anchor node position. 
However, there are several drawbacks of range range-free methods such as DV HOP \cite{dvhop_aps2} systems in the battlefield WSN. The challenge lies in deploying and maintaining the anchor nodes where the terrain is hostile and inaccessible. So the deployments are often done through aerial scattering from airplanes, guided missiles, or balloons which makes the localization problem more complicated. Also, Battlefield environments are constantly changing, with moving troops, vehicles, and obstacles. This can make it difficult to maintain accurate localization information, as anchors and reference points may change or disappear. The radio-based DV-Hop virtual coordinate system doesn't perform well for sparse networks since the correction factor increases with network sparsity. The DV-Hop virtual coordinate system requires the initialization stage to inform each sensor about the hop distance from anchor nodes, which makes the entire network vulnerable to a backtracking attack. Due to the unpredictable behavior of battlefield scenarios, the radio performance may not be stable and reliable which may affect the accuracy of the DV-Hop virtual coordinates system.\\
\begin{figure}[h!]
\centering
\includegraphics[width=0.48\textwidth]{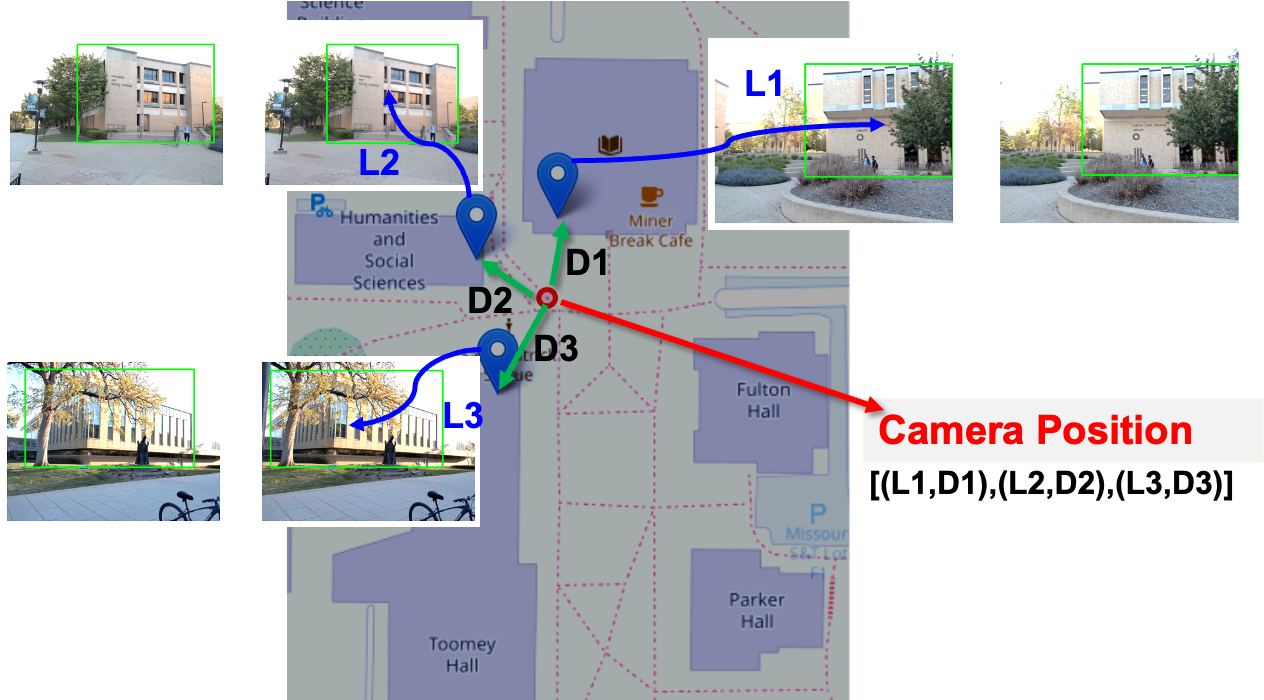}
\caption{Node to Landmark Distance Estimation and Virtual coordinate of Node obtained using Stereo Vision Method} 
\label{depth_overview}
\end{figure}
To that end, we have proposed an alternate strategy of anchor-node deployments in battlefield environments that utilizes naturally existing or man-made landmarks (e.g., buildings, towers, bridges) available within battlefield areas as anchor nodes to obtain the virtual coordinates of mobile nodes as shown in Fig. \ref{depth_overview}. The virtual coordinates are later used with efficient trilateration algorithms for node localization. The virtual coordinate-generating strategy is a multitasking problem that involves a Yolov5-based landmark detection task and efficient stereo vision-based distance estimation from the mobile node to the landmark. The mobile node holding the stereo vision camera on its handheld device captures high-resolution 2D stereo images of the landmark from the real-world scene around itself. Control server(control center that communicates with the mobile client) that either resides within the local device (carried by a moving node) or is established on a remote location, keeps the location information of some pre-defined landmarks that already exist within the battlefield. The Mobile client sends information about recognized landmarks to the control server that keeps the real location of the landmarks. The control center is a computing device that retrieves the location of the recognized landmark calculates the distance from the mobile node to the landmarks using triangulation and obtains the virtual coordinates of the moving node without the use of physically deployed anchors. Also, Our approach utilizes only the vision sensor(camera) to obtain the virtual coordinates of the mobile node.  As per the demand of these two problems, we created two real-world datasets: \textbf{MSTLandmarkv1} and \textbf{MSTLandmarkStereov1}. MSTLandmarkv1 is used to train and validate the landmark detection model while MSTLandmarkStereov1 is used to validate our distance estimation framework. In summary, our key contributions are highlighted below:
    \begin{enumerate}
    \item Creation of custom landmark datasets: MSTLandmarkv1 facilitates model training and MSTLandmarkStereov1 evaluates the distance estimation algorithm.
    \item Landmark-based virtual coordinates system that eliminates reliance on GPS in challenging environments like a battlefield.
    \item YOLOv5 based landmark recognition model with an accuracy of $(0.95 mAP @ 0.5 IoU)$
    \item Improved SGM algorithm integrated with landmark recognition model which ensures efficient and accurate distance estimation.
    \end{enumerate}
This paper is organized as follows: Section II discusses Related Works about different range-free localization methods and distance estimation algorithms. Section III describes our dataset creation approach, and landmark detection model architecture and explains the distance estimation framework in detail. Section IV discusses the Experimental setup, Dataset, model performance metrics, and distance estimation results and analysis. Section V concludes the paper with the future direction of research.

\section{Related Works}
In this section, we will discuss the previous works in the field of Moving object tracking, and localization using active and passive sensing methods.

\subsection{DV-Hop Localization Algorithm}
DV Hop uses the anchor nodes and the vector of minimum hop distance to the anchor nodes to estimate the distance between nodes, which eventually helps to locate the position of an unknown node. DV Hop Virtual Coordinates are achieved using the DV HOP Localization algorithms\cite{dvhop_aps2,dv_hop_Cyclotomic,dvhop_manhattanD,csdvhop}. 
DV-Hop-based range-free localization algorithms are popularly used in wireless sensor networks (WSNs) due to their low-cost hardware requirements for nodes and simplicity in implementation. 
The algorithm estimates the location of unknown nodes based on their hop count to known anchor nodes. This approach considers a sensor deployed with regular sensors with unknown locations and special-type sensors as anchor nodes whose locations are known to the network owners. The main objective of this algorithm is to localize the unknown sensor nodes using the anchor references. The protocol calculates the distances($d_{i,n}$) between anchor nodes and unknown nodes using Eq. \ref{distanceToanchor}.
\begin{equation}
\label{averageHop}
hop_{avg} =  \frac{\sum_{{i}\neq{j}}^{n}\sqrt{(x_i - x_j)^2 + (y_i-y_j^2)}}{\sum_{{i}\neq{j}}^{n}hop_{i,j}}
\end{equation}
\begin{equation}
\label{distanceToanchor}
distance(d_{i,n}) = hop_{avg} * hop_{i,n}
\end{equation}
 
Where, $hop_{avg}$ refers to the average hop distance, while $hop_{i,j}$ is the minimum hop count between unknown anchor node i to j.
Increasing the number of anchor nodes enhances the precision of the distance estimation. However computational and memory overhead also increase by a huge amount. In the worst case, the error rate can be as large as 45 percent of the radio range which could eventually lead to routing failure.
This routing protocol is also error-prone as it uses the estimated location of sensors for routing.


\subsection{DV-Hop Based Geospatial Encoding and Control Message Routing protocol}
DV-Hop-based Method \cite{dvhop_virtualCoords} employs data collection and controlled routing based on DV-Hop virtual coordinates. To route the packet without using GPS-enabled sensors, this algorithm uses a DV-Hop-based virtual coordinate system in which each sensor stores a DV-Hop table that contains information about nearby anchor nodes and the vector of their minimal hop count. 
In this algorithm, the user-defined trajectory is encoded with Geometric shapes such as a segment of Hyperbola or Ellipse. Such Geometric shape can be obtained by combining the anchor node and their hop count relationship. The controlling information in the routing packet contains Geometric shape constraints $e.g. ~C = [A1, A2, A3, a, h3])$ that help the sensor decide whether to rebroadcast the packet or to drop it. If the DV-Hop Information of a sensor satisfies the routing constraint$(h2-h1 = = a ~AND ~ h3 < = r)$ it rebroadcasts otherwise it drops the packet. However,
this approach requires a lot of computational power and time to calculate all possible shape areas and generate the routing constraints. For instance, in a network topology deployed with 80 anchor nodes, it requires approximately three minutes to generate the routing constraint which is not practical for real-time applications.

\subsection{Vehicle Distance Measurement in Autonomous Driving}
In autonomous vehicle systems, detecting surrounding vehicles and measuring their distance in real time is a challenging but very important task. It is key in decision-making for bypassing, changing lanes, or changing the current speed\cite{stereo_vision1}.
Distance Measurement can be done using active and passive methods. Active Method
measures distance by sending a signal (such as laser beams, ultrasound, or radio signal) to an object and capturing the reflected signal using sensors then calculating the Time of Flight. Time of Flight is the time taken by the emitted signal to reach an object and reflect back to the emitter.
\[D = V * ToF/2\]
 Where V is the speed of Signal pulse and ToF is the Time of Flight.\\
Major drawbacks of this method are confusion of echo from previous or the subsequent signal pulse and also the accuracy range is bounded between 1 to 4 meters only. In the Passive method distance is measured by receiving the information about the position of the object using cameras and applying a computer vision technique such as Stereo Matching which calculates the disparity and gives the depth perspective\cite{stereo_vision0}.

\subsection{Vision Based Pose Estimation}
Methods\cite{ref_slam,ref_orbslam,reforbslam2,orb_slam_object_detection}, for robot localization, have gained significant traction in recent years due to their ability to provide accurate and reliable positioning in environments where GPS signals are unavailable or unreliable. These methods utilize image or video data captured by cameras to determine the robot's location within its surroundings. Simultaneous Localization and Mapping (SLAM) technique ~\cite{ref_slam},  allows a mobile robot to simultaneously estimate its position and orientation (pose) within an environment while creating its map. However, it requires a fusion of cameras, lidar, or radar sensors to generate maps. The robot's state is represented by its pose while the map represents specific aspects of the environment such as the location of landmarks or obstacles. 

\subsection{Stereo Matching Algorithms}
Stereo matching algorithms\cite{stereo_vision1,stereo_vision2,stereo_vision3} are crucial in computer vision for depth estimation from stereo image pairs. Semi Global Matching (SGM)\cite{sgm_original} is known for producing accurate depth maps, especially in textured regions. It enforces global consistency in disparity maps, reducing artifacts. However, SGM can be computationally expensive, limiting real-time performance. Performance is sensitive to parameter tuning, making it less robust in different scenarios. Graph Cuts-Based Methods provide global optimization, capturing long-range dependencies in the disparity map and also handle occlusions and depth discontinuities well. Graph cuts can be computationally expensive for large disparity search spaces and performance may degrade in the presence of noise or textureless regions\cite{graph_cut}.CNN based PSMNet \cite{pyramid_matching} 
learn complex features and disparities directly from image data and can handle diverse image content and complex scenes. However, CNN-based approaches often require large amounts of annotated data for training. Also, the interpretability can be challenging, and the reasoning behind disparity predictions may not be transparent.

\section{Our Approach}
Anchor node deployment is a challenging issue in wireless sensor networks that have not been addressed in previous research works. Many real-world applications are unable it is to maintain physical anchors or difficult to set up the DV-Hop table for all devices through anchors’ broadcast.

To mitigate the overhead of maintaining the anchors and updating the DV-Hop table by the sensors, our framework proposes the utilization of geographically existing landmark anchors for for localization of moving nodes. Choosing landmarks as the anchor nodes increases the reliability of the virtual coordinate system as these physical landmarks are more durable than radio-based anchor nodes. We consider the landmark as an anchor and the distance from the vision sensor(say moving node) to the landmarks as their virtual coordinates.


\subsection{System Overview}
The operating environment is within a battlefield where network topology is sparse and dynamic. Sparse networks refer to the WSN where sensors are sparsely distributed while the dynamic means either nodes are moving or nodes may be removed or added to the network dynamically during the network life-cycle. Moving nodes refer to the forces or troop commanders guiding the forces along the battlefield area. We assume that each troop on the battlefield is led by a commander who carries a portable device that is capable of communicating with a server(say Control server). A control server is a powerful computational resource that is capable of performing heavy computational tasks such as running a landmark recognition model for inference and running other algorithms such as disparity and depth calculations. Our framework assumes that dynamic nodes update their distance to nearby landmarks in real time while they change their state and position.

\begin{figure}
\centerline{\includegraphics[width=0.5\textwidth]{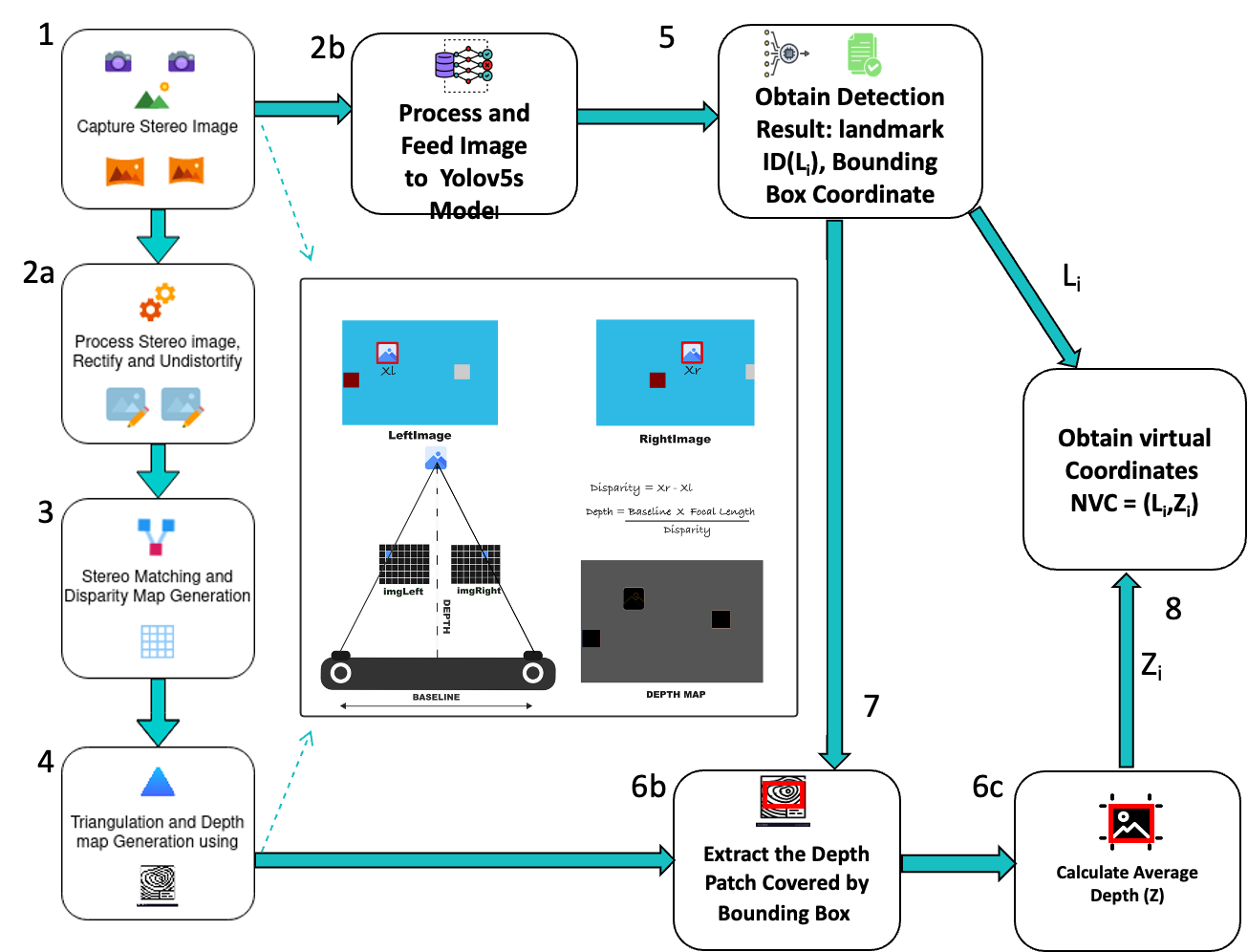}}
\caption{Stereo Vision Method for Depth Measurement}
\label{system_overview}
\end{figure}

We applied the stereo vision method to calculate the distance between an unknown node and the landmark anchor as shown in Fig. \ref{system_overview}. The DV Hop method uses the hop count and the average one-hop distance to the anchor node to estimate the distance from a node to the anchors. But in this approach, we directly calculate the distance to the landmark anchors using the stereo vision method which mitigates the error in distance measurement.

\subsection{Stereo Vision Method}

The stereo vision method uses two cameras to estimate the depth from the camera to the scene elements which can be used to reconstruct the three-dimensional structure of a scene. It works by comparing the image pair of the same scene captured from two identical cameras as shown in Fig. \ref{stereo_rig} separated by a horizontal distance (baseline) to find corresponding points and their disparity in the stereo image pair. The point correspondence problems in our stereo images a solved using the Semi-global Block Matching (SGBM) \cite{sgm_original} that applies the sum of Squared Differences (SSD) as a cost function to ensure the best match.
We implemented the \textbf{Algorithm} \ref{alg_SGBM} and \textbf{Algorithm} \ref{alg_SSD} to generate a disparity map and apply triangulation to eventually obtain the depth map. Consider a stereo pair(ImgLeft, ImgRight) captured from the left and right camera. Due to the horizontal separation of the camera by a baseline(B), an image point u(x,y) in the left frame gets shifted by a distance D in the right frame and the point is found at(x,y) in the right frame. This distance is referred to as the disparity(D) and is given by:
\begin{figure}[h!]
\centering
\includegraphics[width=0.4\textwidth]{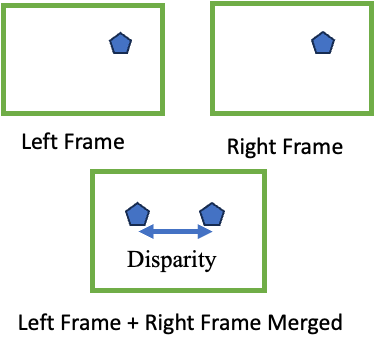}
\caption{Disparity in stereo images} 
\label{fig_disparity}
\end{figure}

A disparity map is generated using Eq. \ref{calc_disp} and the intensity value at each pixel position represents the disparity of that pixel element.
Now, the depth(Z) of each pixel element is calculated based on pixel intensities in the disparity map to generate a depth map using Eq.\ref{triangulation}:
\begin{equation}
D = X_{ImgLeft}- X_{ImgRight}
\label{calc_disp}
\end{equation}

\begin{equation}
Z = f_{pixel} * B / D
\label{triangulation}
\end{equation}

where f is the focal length of the camera.
Eq.2 indicates that disparity is inversely proportional to depth. This means that the disparity between corresponding image points increases when the scene point approaches nearer to the camera. Disparity calculation is a crucial step in stereo vision algorithms, as it provides the necessary information to estimate the depth of objects in the scene.

\begin{algorithm}[H]
\label{sgm_algo}
\caption{Semi-Global Block Matching Algorithm}
\label{alg_SGBM}
\begin{algorithmic}[1]
\Require Stereo images $I_l$ and $I_r$
\Require Block size $blockSize$
\Require Disparity range $minD \ldots maxD$
\Require Discount factor $\alpha$
\Ensure Disparity map $D$

\State Initialize disparity map $D$ to zero
\State Initialize cost matrices $C_{l}$ and $C_{r}$ to zero
\ForAll {$d$ in $minD \ldots maxD$}
\State Compute the sum of squared differences (SSD) between the block in $I_l$ and the corresponding block in $I_r$ with disparity $d$
\State Compute the cost function $C_d$ based on the SSD and discount factor $\alpha$
\State Update the cost matrices $C_{l}$ and $C_{r}$ based on the cost function $C_d$
\State Find the disparity value $d$ that minimizes the cost function $C_d$ at each pixel in the block
\State Update the disparity map $D$ by selecting the disparity value that minimizes the cost function $C_d$
\EndFor
\end{algorithmic}
\end{algorithm}

\begin{algorithm}[H]
\label{ssd_algo}
\caption{Sum of Squared Differences (SSD)}
\label{alg_SSD}

\begin{algorithmic}[1]
\Require Reference image $I_r$
\Require Template image $I_t$
\Require Displacement vector $d$

\Ensure Matching score $S$

\State Initialize matching score $S$ to zero
 \ForAll {$p$ in ${I_t}$}
    \State Compute the intensity difference between pixel $p$ in $I_t$ and the corresponding pixel $p+d$ in ${I_r}$
    \State Square the intensity difference and add it to the matching score $S$
    \State Normalize the matching score $S$ by dividing it by the number of pixels in $I_t$
 \EndFor
\end{algorithmic}
\end{algorithm}

\subsection{Stereo Dataset Creation Framework}
\begin{figure}[htbp]
    \centerline{\includegraphics[width=0.45\textwidth]{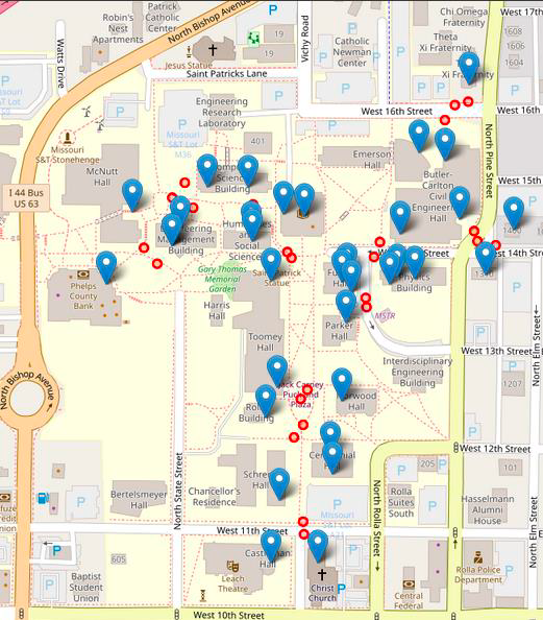}}
    \caption{Area Selected For Landmark Dataset Creation. The blue pin  indicates the landmarks and the red solid circle indicates the position of the stereo camera while capturing the images.}
    \label{selected_area}
    \end{figure}

    We created real-world landmark stereo datasets to test the validity of our distance estimation framework. MSTLandmarkStereov1 is used for stereo vision-based distance measurement and testing. To obtain a custom landmark stereo dataset, we built a stereo camera setup that captures and stores stereo images of the landmarks and the location from where the image pairs were captured. 
    We selected an area shown in Fig.\ref{selected_area}, rich with prominent landmarks(buildings) within Missouri University of Science and Technology premises, and captured around 1000 stereo images of 34 different landmarks. All the steps involved in the dataset creation framework are discussed below in detail. 
\subsubsection{Building a custom Stereo camera setup}
\begin{figure}[htbp]
    \centerline{\includegraphics[width=0.5\textwidth]{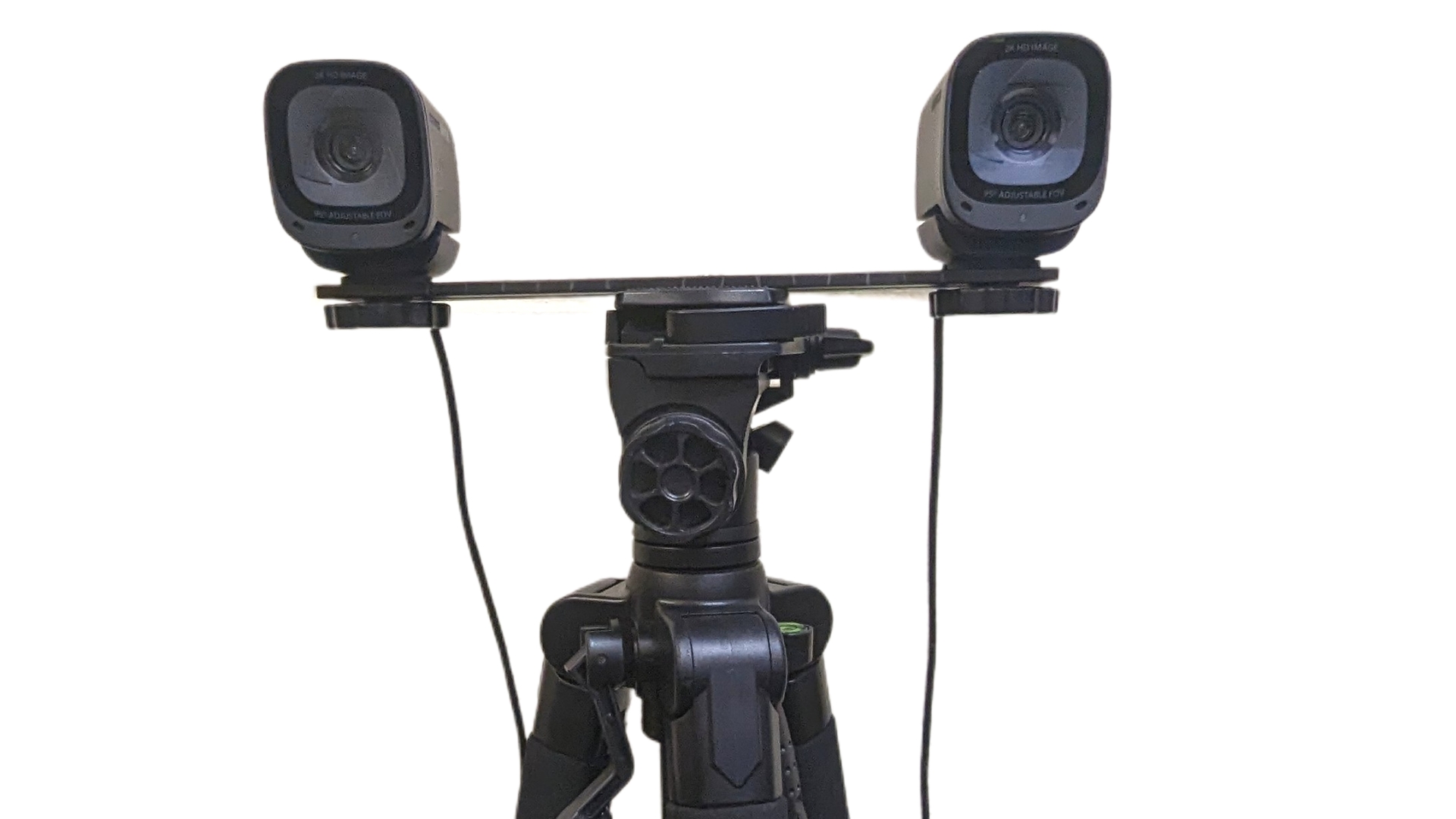}}
    \caption{Custom Stereo Vision Setup}
    \label{stereo_rig}
    \end{figure}

    we built our own stereo rig setup for image capture tasks. We used two identical 2K-resolution (2560*1440px @30 fps) cameras as shown in Fig.\ref{stereo_rig}  having an adjustable field of view (FoV) of [50, 70, 95] degrees. We arranged the camera on the long metal bar with an adjustable baseline(B) from $10-40 ~ cm$. The baseline indicates the horizontal separation  of the stereo cameras. The metal bar attached to two cameras was mounted to a height-adjustable tripod stand.
    
    \subsubsection{Camera Calibration}
    Stereo camera calibration is essential for achieving accurate depth perception, image rectification, and lens distortion correction. Stereo vision relies on the disparity between corresponding points in the left and right images captured by the stereo camera system. Calibrating the cameras helps determine the intrinsic and extrinsic parameters, which are essential for accurate depth perception. Calibration is necessary to rectify the stereo images, ensuring that corresponding points in the left and right images lie on the same rows. Rectification simplifies the stereo-matching process and facilitates the computation of disparities. Cameras may introduce distortions in the images due to lens imperfections. Calibration helps correct these distortions, ensuring that the images accurately represent the scene geometry. Camera calibration refers to both intrinsic and extrinsic calibration. Through the intrinsic calibration, we determine the image center, and focal length, while the extrinsic calibration determines the $3D$ positions of the cameras.

\begin{figure}[h!]
    \centerline{\includegraphics[width=0.5\textwidth]{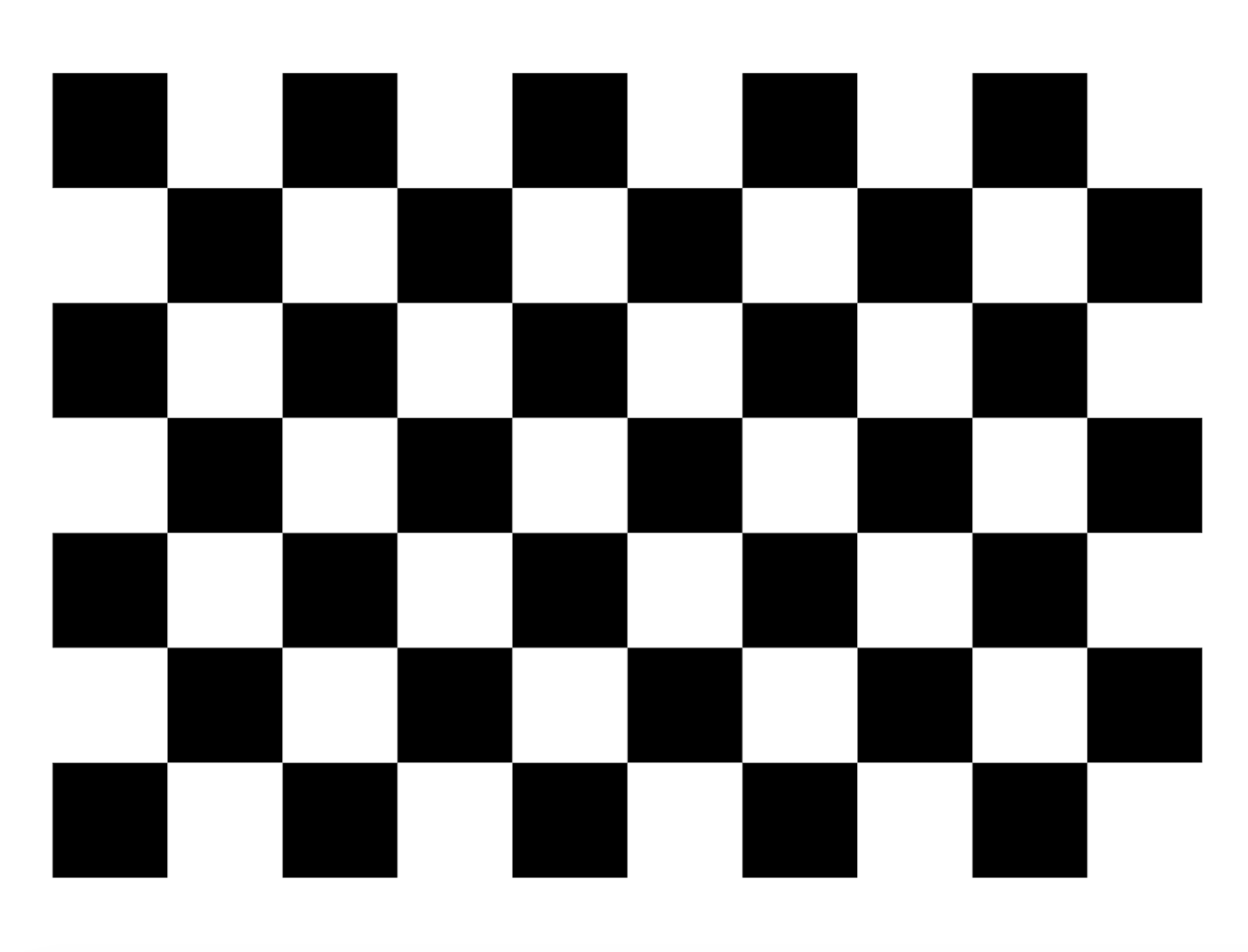}}
    \caption{9*6 Checker Board Pattern Used for Camera Calibration}
    \label{checker_pattern}
    \end{figure}

    To perform the camera calibration task we used the checkerboard method available in the OpenCV library. For this purpose, we obtained the $9*6$ checkerboard as shown in the Fig. \ref{checker_pattern}.

    \subsubsection{Essential and Fundamental Matrices}
    As a result of camera calibration, we obtain camera intrinsic and extrinsic parameters. we have used the $stereoCalibrate()$ method from openCV to find the transformation between the stereo camera pair, the essential matrix (E), and the Fundamental matrix (F).\\
    
    The fundamental matrix is a $3x3$ matrix that encapsulates intrinsic parameters such as  Focal Length (fx, fy) and Principal Point (cx, cy) of two calibrated cameras in a stereo vision system.
     \[
        F = 
        \begin{bmatrix}
        \label{fundamental_matrix}
        f_x & 0 & c_x\\
        0 & f_y & c_y\\
        0 & 0 & 1
        \end{bmatrix}\]
    The essential matrix is also a 4x4 matrix, that captures the extrinsic (calibration) parameters of the cameras along with the geometric information. It relates the motion and geometry between two calibrated cameras and provides a compact representation of the relative pose (rotation and translation) between the cameras. The essential matrix can be decomposed into rotation and translation matrices, allowing the extraction of the relative pose between the cameras.

    \[E = 
         \begin{bmatrix}
         \label{essential_matrix}
        r_{11} & r_{12} & r_{13} & t_x\\
        r_{21} & r_{22} & r_{23} & t_y\\
        r_{31} & r_{32} & r_{33} & t_z\\
        0    & 0    & 0    & 1
        \end{bmatrix}
       \]
\subsection{Landmark Recognition using Yolov5}
We used Deep Convolutional Neural Networks \(CNN\) to detect and localize the landmark anchors within the image. Since we rely on the camera sensors to estimate the distance to the landmark and captured stereo images for the localization of the node, we chose to apply recognition and localization of the landmark anchor using the YOLOv5 based object detection model\cite{yolov5_github}. The ability to detect objects in a single pass and at different scales makes YOLO well-suited for identifying larger objects efficiently with fewer localization errors. $YOLOv5$ takes a single-stage approach to object detection, handling all tasks \(feature extraction, prediction, and detection\) in one pass. YoLov5 Network contains three key components: Backbone Head and Neck. It uses a $CSP-Darknet53$ backbone which efficiently extracts relevant features from the input image. The neck component, combining spatial pyramid pooling $(SPP)$ and PANet, utilizes features from various backbone levels for a richer representation. The head component predicts bounding boxes and class probabilities for objects in the image. The SPPF layer boosts network speed by pooling features of different scales into a fixed-size feature map. Each convolution utilizes batch normalization and SiLU activation for optimized performance. By combining SPP and PANet, YOLOv5's neck effectively combines features from different levels, enhancing representation richness. YOLOv5's single-stage architecture and efficient component design make it a powerful tool for object detection tasks. 

\subsubsection{Creation of Real-World Landmark Dataset}
To meet the demand of our Yolov5-based landmark recognition task we created another real-world dataset that comprises around ~7500 instances of 34 different landmarks and named it as \textbf{MSTLandmarkv1} dataset. The images in the dataset were captured from the selected area within Missouri University of Science and Technology as shown in Fig.\ref{selected_area} using high-resolution monocular cameras and stereo cameras and then merged to a single directory. The landmark images were carefully captured in different lighting conditions from different angles and orientations to achieve different perspective views of the landmarks. The landmark images were manually annotated using the roboflow.ai platform and exported to Yolov5 format. The dataset is split into training, validation, and testing sets in the ratio of 70:20:10. 
\begin{figure}[h!]
\centering
\includegraphics[width=0.48\textwidth]{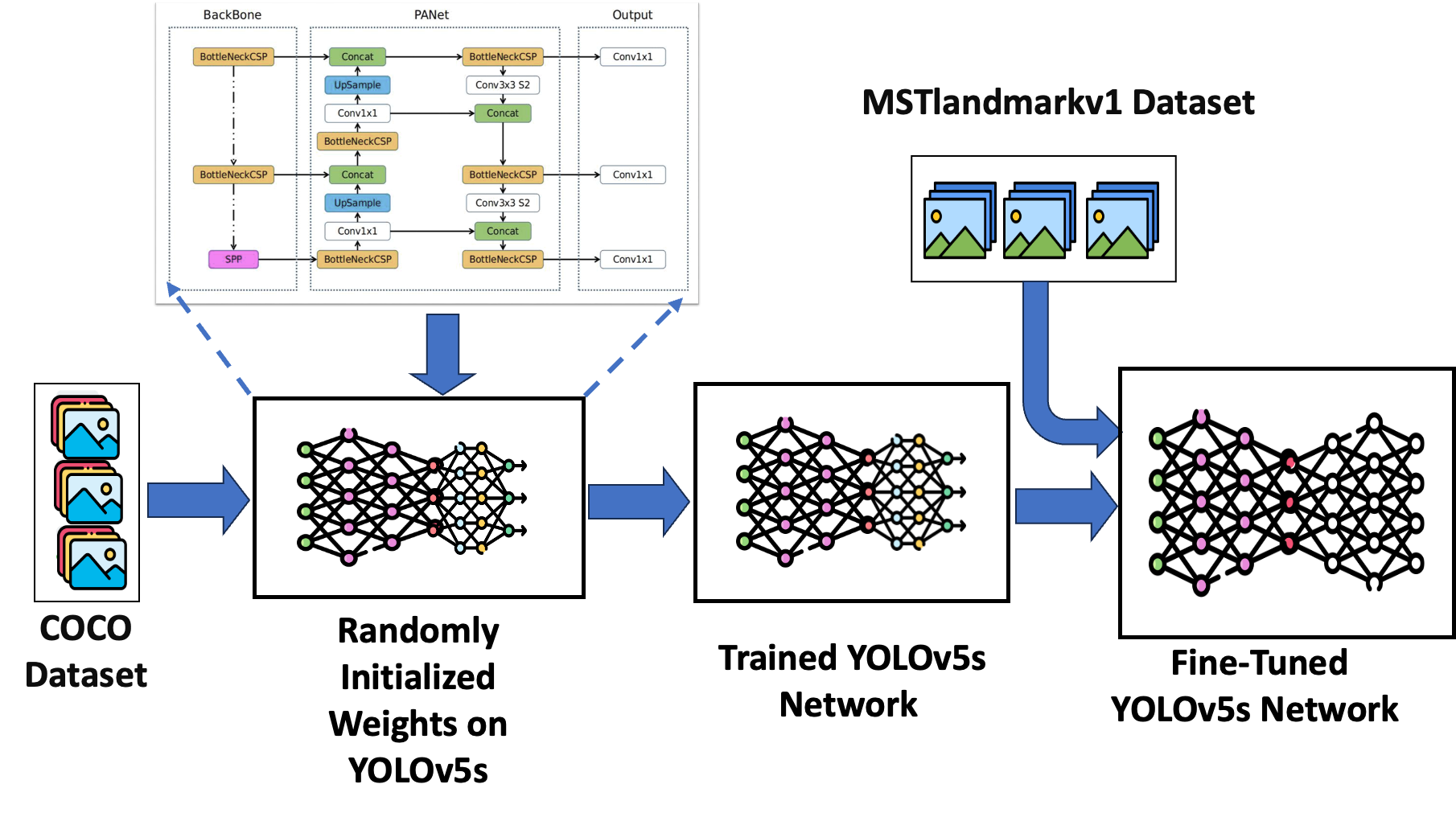}
\caption{Fine Tuning YOLOv5s with MSTLandmarkv1 Dataset} 
\label{fig_yolo_network}
\end{figure}
We selected pre-trained YOLOv5s and re-trained the model with our landmark dataset to fit our recognition task. The model detects and recognizes the landmark in real-time with an average speed of $0.3ms$ for pre-process, $7.1ms$ inference, and $1.3ms$ post-process per image. The detection result gives the bounding box coordinates of the detected landmark, which are used to extract the depth patch of the detected landmarks.
\subsection{Landmark Distance Estimation}
Our algorithm performs several steps and procedures to calculate the distance to landmark anchors in real time. A stereo vision camera captures the images of the scene within its field of view. We consider that captured images contain a visible landmark on it. The algorithm detects and identifies the landmark object from the captured Stereo Images and sends the results to the local server within the mobile device. The result  of landmark detection would look like below:

\texttt{
    {\\
      $``x": 430,$\\
      $``y": 202$,\\
      $``width": 420,$\\
      $``height": 297,$\\
      $``confidence": 0.9567493200302124,$\\
      $``class": ``Curtis ~Laws~ Wilson~ Library 1",$\\
      $``classId": 11,$\\
      $``imagePath": ``img8.png",$\\
      $``predictionType": ``ObjectDetectionModel"$\\
    }
}
\begin{figure}[h]
\includegraphics[width=0.45\textwidth]{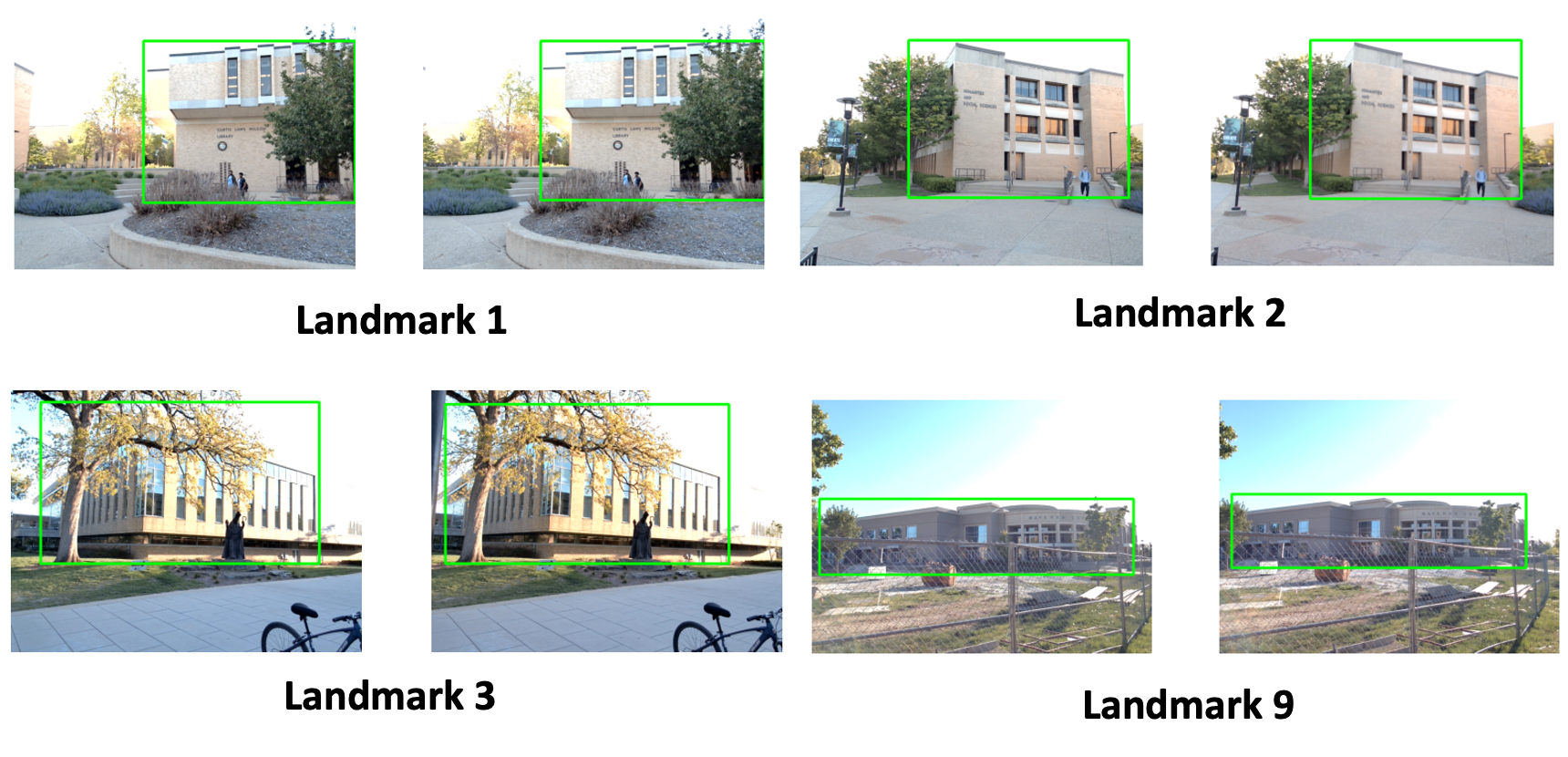}
\centering
\caption{Example of Landmark Detection with Bounding Box} 
\label{detected_landmark}
\end{figure}

\subsubsection{Depth Patch Extraction:}
We used an improved SGM \cite{sgm_improved} algorithm to generate the disparity map of stereo image frames. The depth map is then generated using camera parameters (focal length and baseline) and disparity values corresponding to each pixel position in the disparity map and the triangulation formula shown in Fig \ref{system_overview}. The bounding box coordinates obtained from the landmark detection model are used  to crop and extract the depth patch of the landmark from the depth map. The depth patch comprises, the depth features of the detected landmark, which is further aggregated to calculate the average depth to the landmark anchor. This step is particularly important in obtaining a more precise distance calculation to the landmark.

\subsubsection{Landmark Distance and Virtual Coordinate Estimation:}
    As an input, we used stereo image sets containing landmarks and objects captured using our own calibrated stereo vision camera. we processed, normalized, and resized each stereo image to 640*480 resolution before feeding it into the landmark detection model and stereo-matching algorithm. \\
    
    The steps involved in distance calculation to obtain the virtual coordinates of a node are summarized in the \textbf{Algorithm \ref{alg_dist_estm}}.

 \begin{algorithm}[H]
\caption{Landmark Distance and Virtual Coordinate Estimation}
\label{alg_dist_estm}
    \begin{algorithmic}[1]
        \Require: Stereo images
        \Ensure:  Distance to landmark  \\
        \textbf{procedure} MEASUREDISTANCE\textit{(imgL,imgR)}:
        \ForAll {$l \in L$}
            \State Capture Stereo image set (imgL,imgR) then rectify(imgL,imgR) and undistortrectify(imgL,imgR) using calibration parameters.
            
            \State {Detect and localize the LANDMARK using Yolov5s Model}
            \State Calculate disparity in (imgL,imgR) using \textbf{Algorithm \ref{alg_SGBM}} and \textbf{Algorithm \ref{alg_SSD}} and generate Disparity map.
            \State Generate Depth map using Disparity map from Step (5) and apply triangulation using calibration parameters.
            \State Obtain the depth patch from the depth map obtained in Step (6) utilizing the bounding box coordinates.
            \State Aggregate and Compute the average depth from the depth patch.
             \State  Store landmark id and depth as the Node Virtual Coordinate: $nvc = (L_{id}, L_{dist})$
        \EndFor
        \State Repeat Steps (1) to (7) for at least 3 landmarks in the trilateration set then stop.
    \end{algorithmic} 
\end{algorithm} 

A tuple of  Landmark id $l_{i}$ and the distance to landmark $d_{i}$ are stored as virtual coordinate $(li,di)$ of a moving node for a particular instance. The virtual coordinates of the landmarks are later used to localize the position of the moving node. 
\section{Experiments}
\subsection{Experimental Setup}
To test the validity of our landmark distance estimation framework, the experiment was performed on Alienware Aurora R12 System with 11th Gen Intel Core i7 CPU, 32 GiB Memory, and NVIDIA GeForce RTX 3070 GPU using Python 3.10 on PyCharm 2022.1 (Edu) IDE. The landmark recognition model Yolov5 was trained on the Google Colab environment with Python-3.10.12 and torch-2.1.0+cu118, utilizing an NVIDIA Tesla T4 GPU with 16GiB memory.
\subsection{Dataset}
We used the following datasets for our experiments.
\subsubsection{MSTlandmarkv1 Dataset:}
\begin{figure}[h!]
    \centering
     \includegraphics[width=0.5\textwidth]{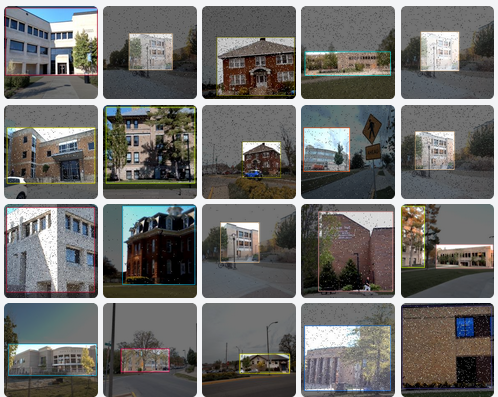}
     \caption{Example landmarks MSTlandmarkv1 dataset}
     \label{landmark_sample_data}
\end{figure}

\begin{figure}[h!]
    \centering
     \includegraphics[width=0.48\textwidth]{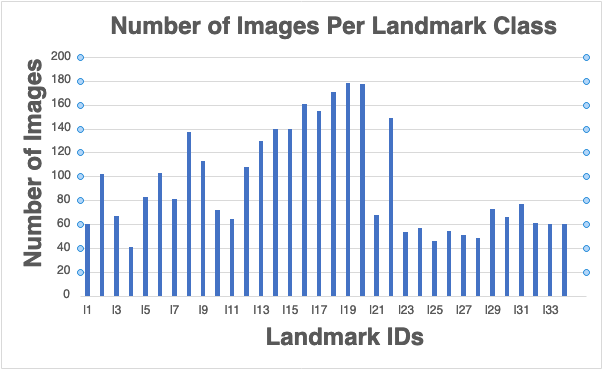}
     \caption{Histogram of Number of Landmarks image Classwise in MSTlandmarkv1 dataset}
     \label{image_dist_classwise}
\end{figure}

We created the MSTlandmarkv1 dataset to train our landmark recognition model comprising around ${4000}$ images of $34$ different landmark instances as shown in Fig.\ref{landmark_sample_data} which were labeled manually using roboflow.ai\cite{roboflow_platform}. The image distribution per class of landmark is shown in Fig.\ref{image_dist_classwise}. The images were captured using a camera with $2K$-resolution $(2560*1440px)$. The distribution of images per landmark class in the MSTlandmarkv1 dataset is shown in Fig.\ref{landmark_stereo_examples}. The dataset was split to train($70$ percent), validation($20$ percent), and test($10$ percent) sets. To prevent overfitting problems and to improve the model robustness we did image augmentation on the training portion of the dataset. We performed image color jittering by adjusting brightness between $-25$ percent and $+25$ percent, bounding box rotation between -15° and +15°,  and bounding box noise addition up to $5$ percent of the pixels. Finally, we exported the dataset in Yolo V5 Pytorch format comprising $7547$ images including train, validation, and test sets.


\subsubsection{MSTlandmarkStereov1 Dataset:}
\begin{figure}
    \centering
     \includegraphics[width=0.48\textwidth]{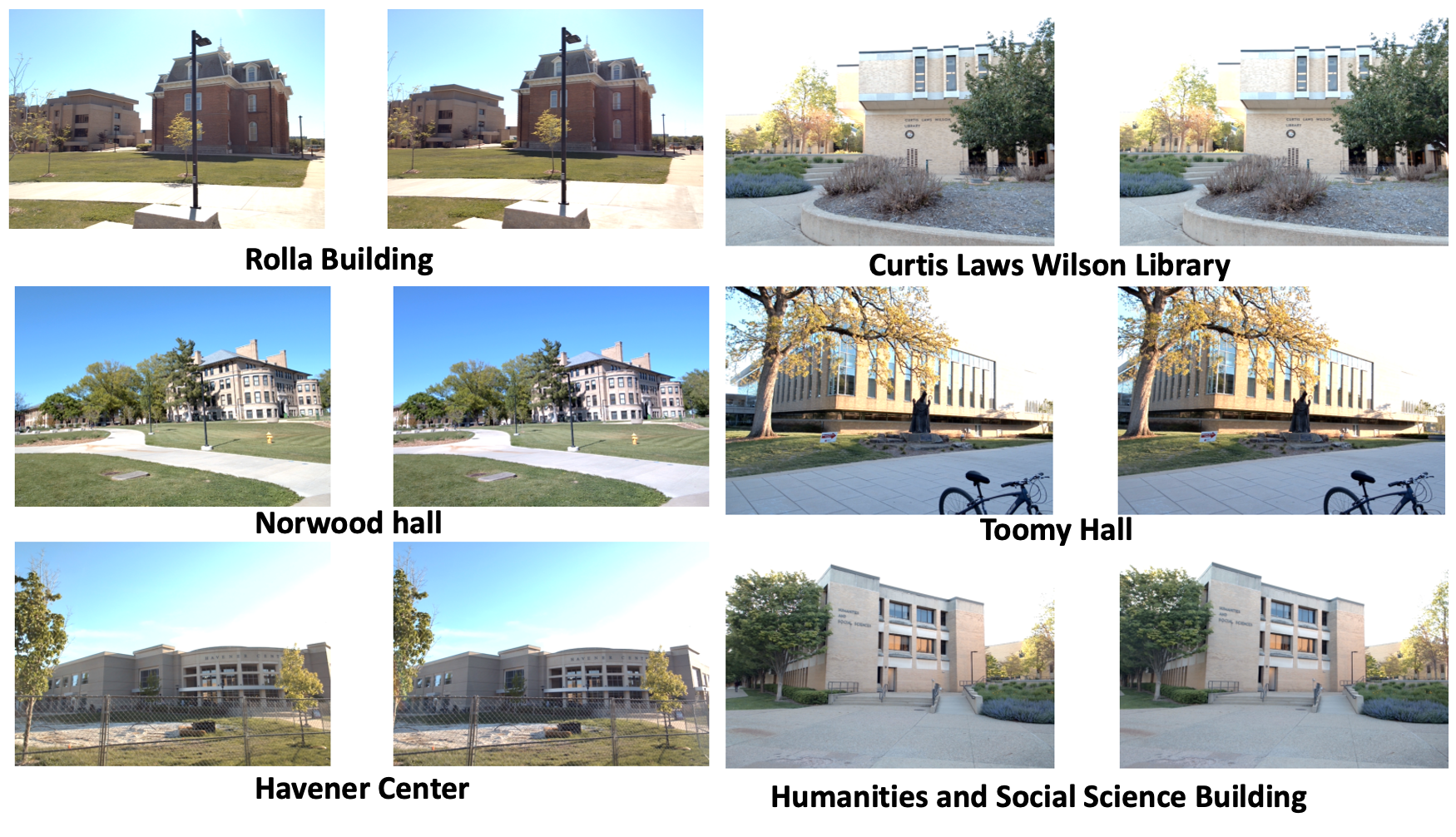}
     \caption{Example Landmark MSTlandmarkStereov1 images}
    \label{landmark_stereo_examples}
\end{figure}

\begin{figure}
    \centering
     \includegraphics[width=0.4\textwidth]{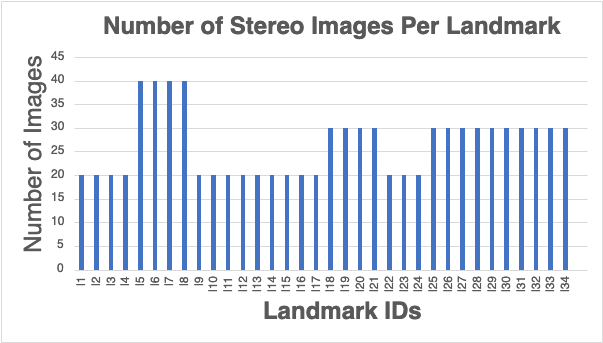}
     \caption{Histogram of Number of Landmark Stereo images per Class in MSTlandmarkStereov1 dataset}
     \label{stereo_image_dist}
\end{figure}

We created another real-world landmark stereo dataset shown in Fig.\ref{landmark_stereo_examples} using our stereo image capture framework. We used two identical cameras of $(2560*1440px~@~30~fps)$ resolution as shown in Fig. \ref{stereo_rig} with adjustable field of view (FoV). The cameras were arranged on the long metal bar which allowed us to adjust the baseline(B) from $10-40 ~ centimeters$. The distribution of stereo image samples per landmark class is shown in Fig.\ref{stereo_image_dist}. This dataset is used to perform distance estimation experiments from unknown nodes to the landmark anchors using the distance estimation \textbf{Algorithm \ref{alg_dist_estm}}.
\subsection{Trainnig Model}
We took the model $YOLOv5$ from Ultralytics\cite{yolov5_github} and trained it using our \textbf{MSTlandmarkv1} dataset. We adjusted the number of epochs and modified the upper layers of the network to learn and classify $34$ different classes of landmarks present in our dataset. Training a model for $50$ epochs took $0.971$ hours. 

The important hyperparameters used in training the model are shown in Table \ref{training_hperparameters}.

\begin{table}[h!]
\centering
\begin{tabular}{|l|l|}
\hline
\textbf{Hyperparameters} & \textbf{Values} \\ \hline
Optimizer                & SGD             \\ \hline
Learning Rate initial (lr0) & 0.01          \\ \hline
Learning Rate Final (lrf) & 0.01            \\ \hline
Warmup Momentum          & 0.8             \\ \hline
Momentum(Beta1)          & 0.937           \\ \hline
Weight Decay             & 0.0005           \\ \hline
Warmup Epochs            & 3.0               \\ \hline
Epoch                    & 30              \\ \hline
Batch Size               & 16              \\ \hline
\end{tabular}
\vspace{0.1in}
\caption{Hyperparameters used to Train the model}
\label{training_hperparameters}
\end{table}
\subsubsection{Training and Validation Results}
\begin{figure}[h!]
\centering
\includegraphics[width=0.5\textwidth]{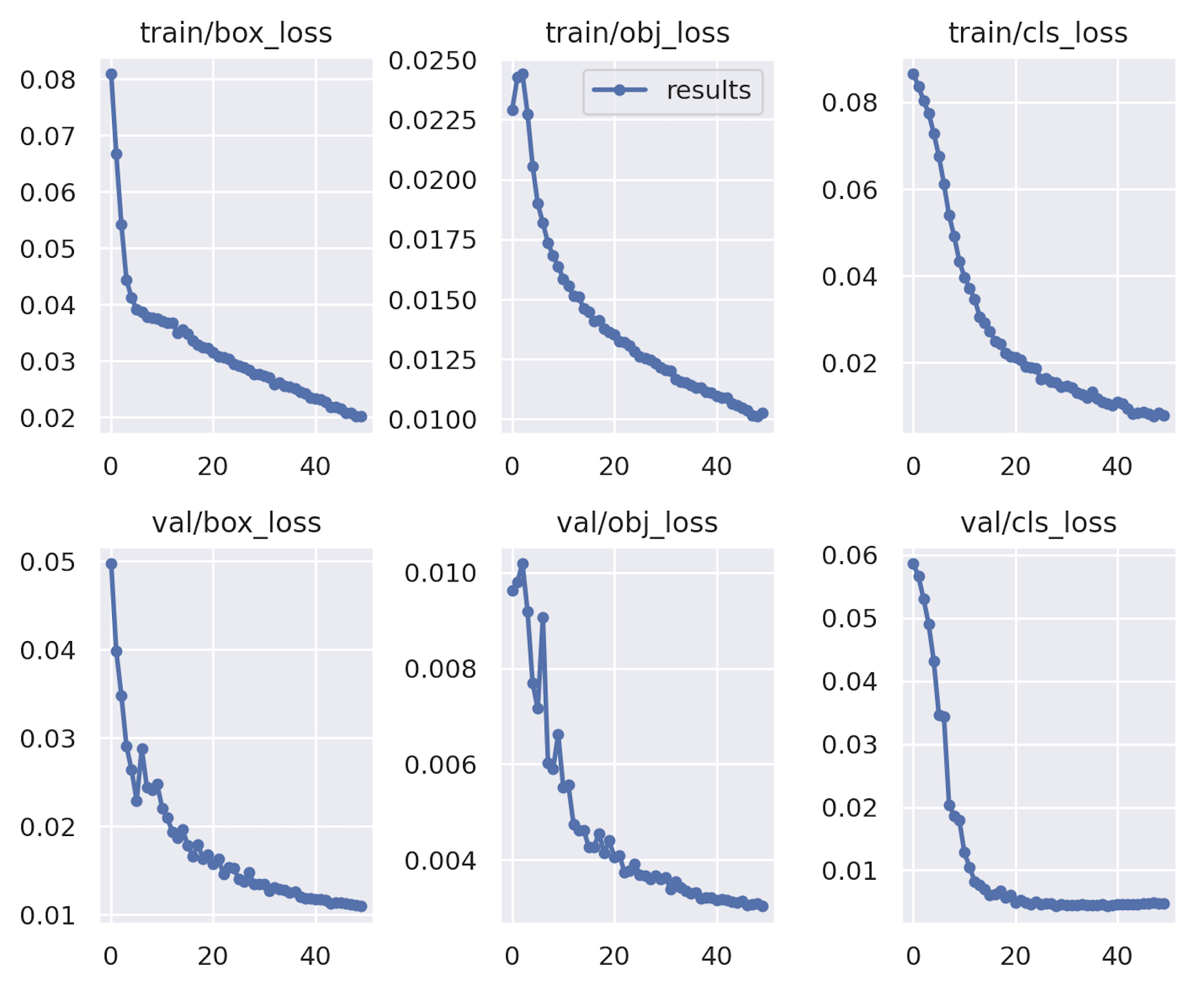}
\caption{Graph showing box loss, objectness loss, classification loss over the training epochs for the training and validation} 
\label{fig_train_result1}
\end{figure}

\begin{figure}[h!]
\centering
\includegraphics[width=0.5\textwidth]{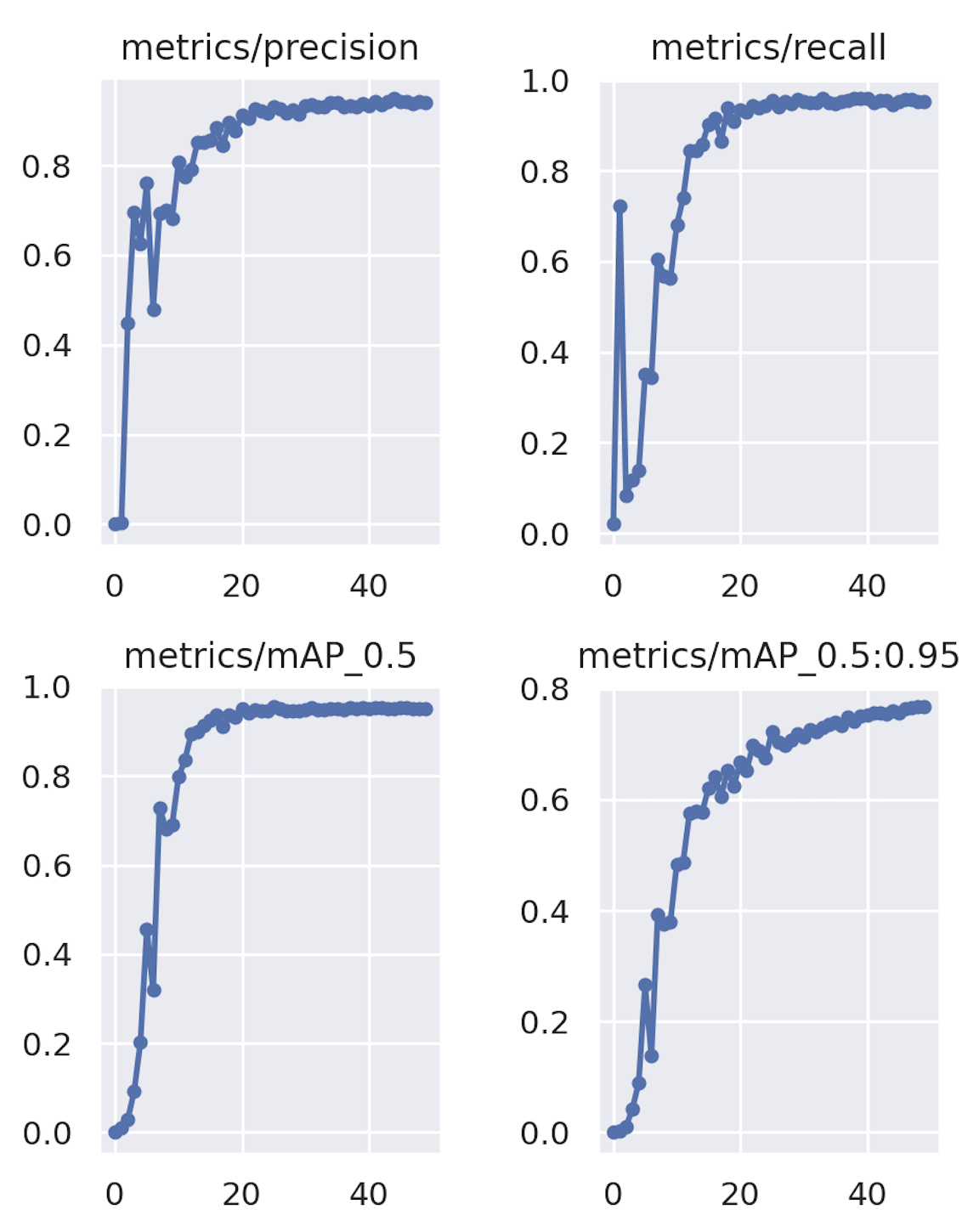}
\caption{Graph showing precision, recall, and mean average precision (mAP) over the training epochs for the training and validation} 
\label{fig_train_result2}
\end{figure}

\begin{figure}
\centering
\includegraphics[width=0.5\textwidth]{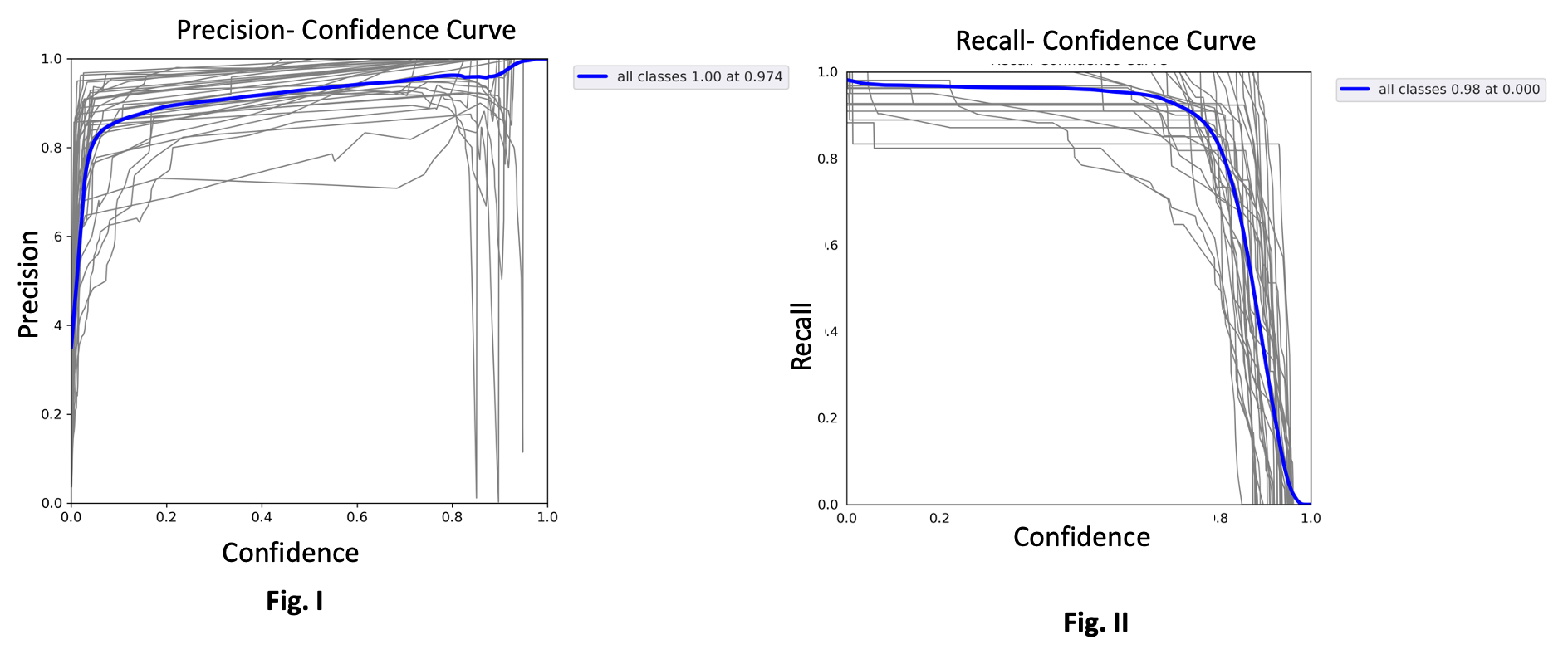}
\caption{Prediction Metrices Plot:Fig.1 refers to  Precision-Confidence Curve and Fig.II refers to Recall-Confidence Curve} 
\label{curve2}
\end{figure}

Fig. \ref{fig_train_result1} represents graphs showing how box loss, objectness loss, and classification loss are improving over the training Epochs. Fig.\ref{fig_train_result2} represents the model performance scores (Precision, Recall, and mAP) as a result of training and validation. Box loss represents how well the predicted bounding box aligns with the ground truth bounding box. Objectness Loss typically represents the likelihood that an object is present in a proposed region. This helps the model distinguish between regions containing objects and those that do not. Classification loss measures how well the model predicts the class of the detected object.
The model's performance improved significantly in terms of precision, recall, and mean average precision (mAP) until it reached a stable point after about 40 epochs as shown in Fig.\ref{fig_train_result1} and Fig. \ref{fig_train_result2}. The box, objectness, and classification losses for the validation data also decreased rapidly until around epoch $40$. Therefore, to optimize model selection, we implemented early stopping, choosing the weights corresponding to the peak performance as the best model. Our best model shows an optimal Box loss precision(Pr) of $0.94$, a Recall (R) of $0.953$, and achieved a mean Average Precision $(mAP~@~0.5~IoU)$ score of $0.95$ and $mAP~@~[0.5:0.95] ~ IoU$ of $0.767$ for all classes as shown in Fig.\ref{fig_train_result2} and Fig.\ref{curve2}.

\subsubsection{Test Result}
\begin{figure}[h!]
\centering
\includegraphics[width=0.5\textwidth]{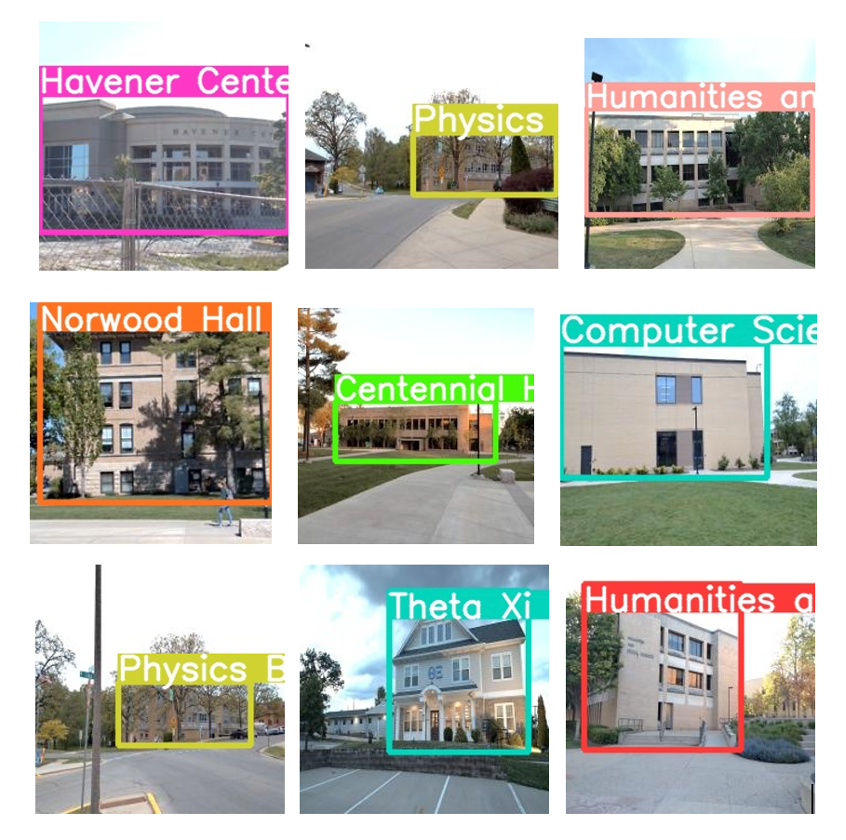}
\caption{Prediction results of different landmark images from test datasets which were truly predicted} 
\label{test_results}
\end{figure}

The model detects and recognizes the landmark in the test sets with an average inference speed of speed 7.1ms, taking 0.3ms for pre-processing, and 1.3ms for post-processing per image at shape (1, 3, 416, 416). The prediction result of some example landmarks from test sets are shown in Fig. \ref{test_results} with bounding boxes and class labels.

\subsection{Distance Measurement Using Stereo Vision}
\begin{figure}[h!]
\includegraphics[width=0.50\textwidth]{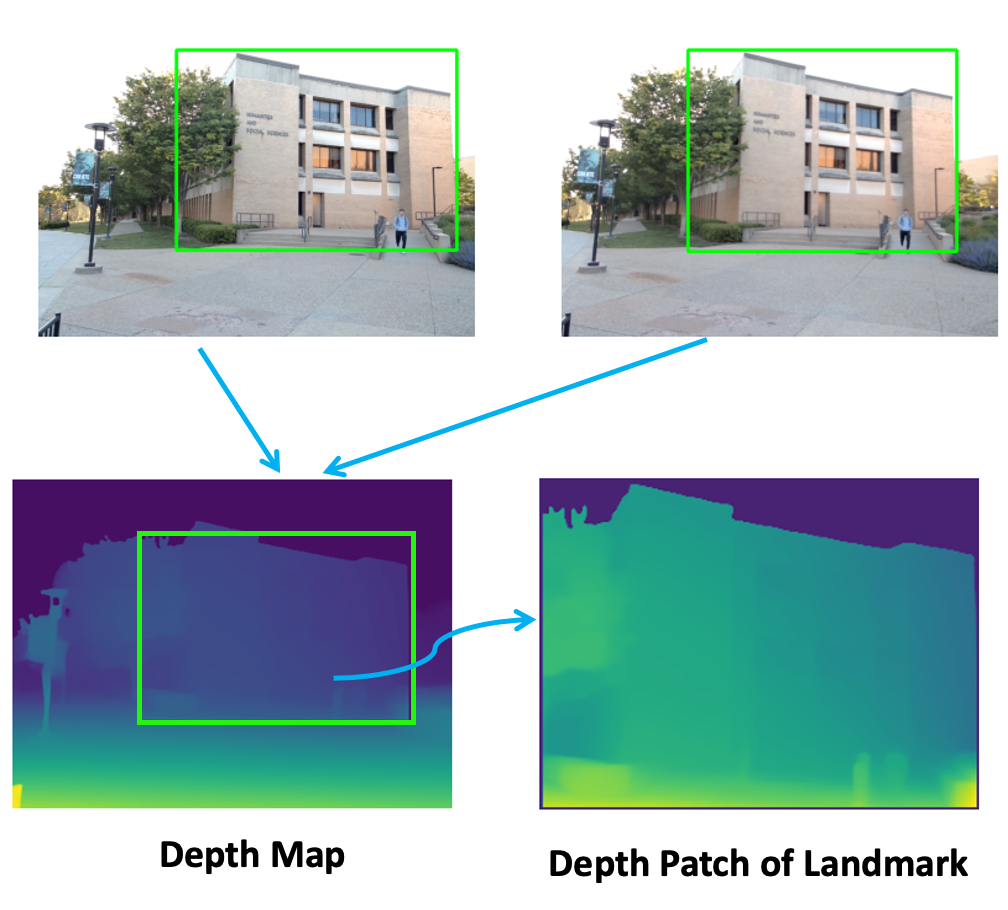}
\centering
\caption{Showing Depth Patch Extraction Using Bounding Box Coorinates} 
\label{depth_extract}
\end{figure}

For distance measurement from the unknown node position to the landmark anchors, we applied the \textbf{Algorithm \ref{alg_SSD}} for disparity map generation and then applied triangulation using Eq.\ref{triangulation} to obtain the depth map.
The depth patch containing the landmark anchor is cropped and extracted using bounding box coordinates obtained from landmark recognition results as shown in Fig. \ref{depth_extract}, which is aggregated to obtain an estimated distance from a node carrying the stereo camera.

The distance measurement test was performed for 8 different trilateration sets each containing at least 3 landmark anchors. In each trilateration experiment, the distance from a camera position to the landmarks was calculated. The calculated(observed) distance was compared to the actual distance(ground truth) present in the \textbf{MSTlandmarkstereov1}. The actual distance from a node to the landmark was calculated using the great-circle distance method \ref{great_circle} while creating the MSTlandmarkstereov1 dataset. Great-circle distance computes the distance between two geographic coordinates(landmark anchor position and node position in our case) on the surface of the earth. The actual node position was captured and stored while creating the \textbf{MSTlandmarkstereov1} dataset.\\

\begin{figure}[h!]
\centering
\includegraphics[width=0.5\textwidth]{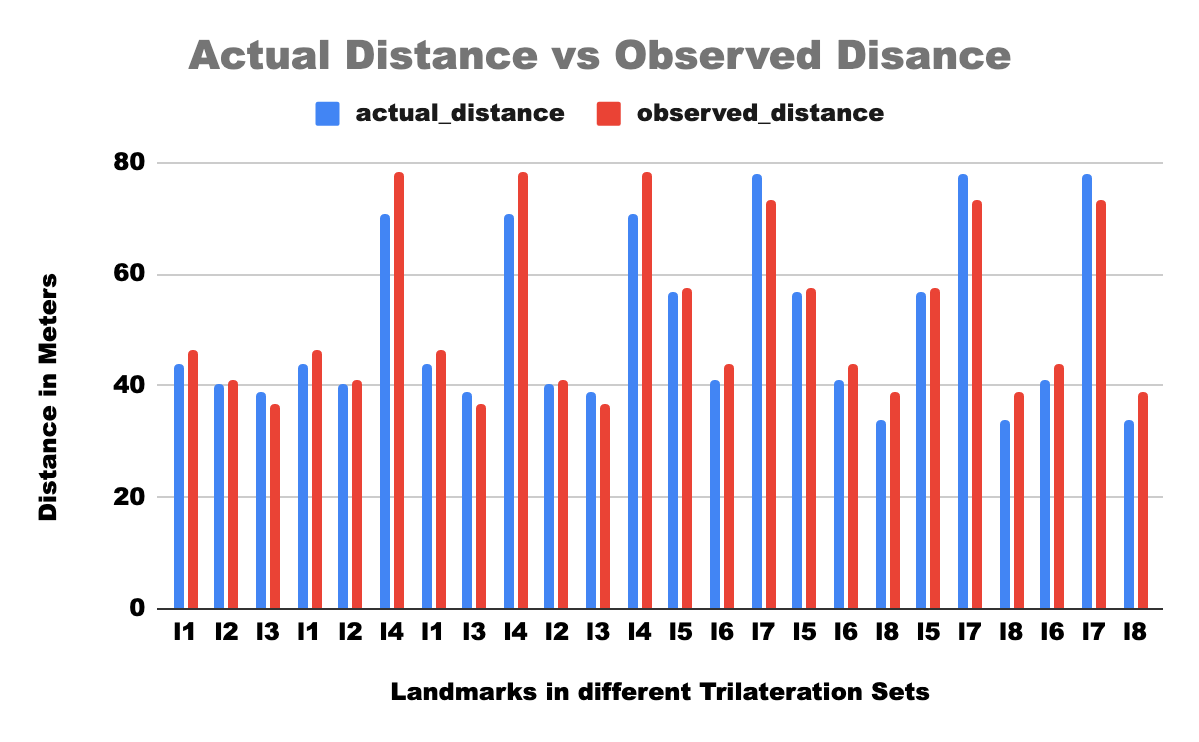}
\caption{Column chart showing actual distance and observed distance to different landmarks from a node position in different trilateration sets. The blue column indicates the actual distance and the red column indicates the observed distance using the stereo vision method.} 
\label{distance_graph}
\end{figure}

\begin{figure}[h!]
\centering
\includegraphics[width=0.5\textwidth]{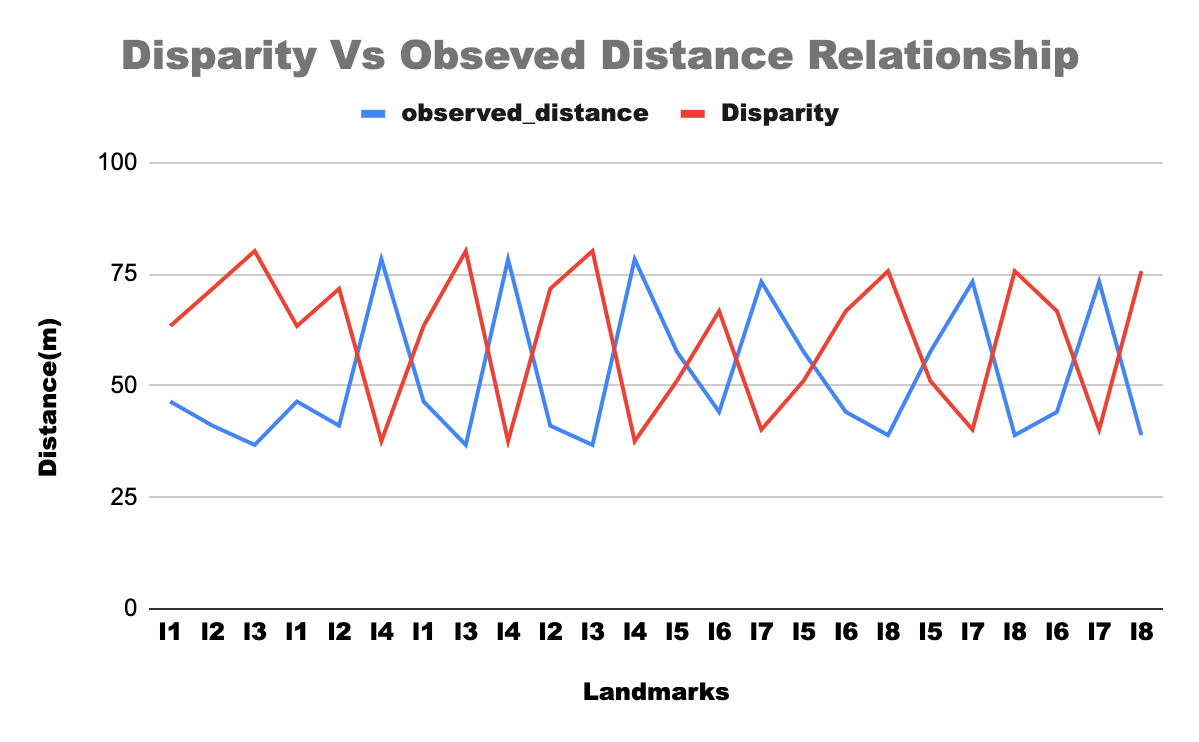}
\caption{Showing the Reciprocal Relationship between Average Observed Distance and Average Disparity Measurement of the Landmarks in MSTLandmarkStereov1 Dataset} 
\label{distance_vs_disparity}
\end{figure}

The results of the distance estimation are  shown in Fig..\ref{distance_graph} and the result statistics of observed versus actual distances are shown in Table \ref{distance_table}. Our approach achieved a Root mean square Error (RMSE) of 1.79m within the range of 33-78 meters. The absolute error fluctuates between 0.49 meters to 7.3 meters with a mean of 2.36 and a Standard Deviation of 2.20.\\
To validate our estimated distance as defined by Eqn. \ref{triangulation}, we evaluated the relationship between the average observed disparities of sample stereo images and the observed distance in our result. Fig. \ref{distance_vs_disparity} shows that when disparity has increased, the estimated distance has been decreased which means that objects with larger disparities appear closer to the camera and vice versa.  

\begin{table}[h!]
\centering
\begin{tabular}{@{}ll@{}}
\toprule
\rowcolor[HTML]{FFFFFF} 
{\color[HTML]{333333} \textbf{Metrices}}          & {\color[HTML]{333333} \textbf{Values(M)}}         \\ \midrule
\rowcolor[HTML]{FFFFFF} 
\multicolumn{1}{|l|}{\cellcolor[HTML]{FFFFFF}MIN} & \multicolumn{1}{l|}{\cellcolor[HTML]{FFFFFF}0.49} \\ \midrule
\multicolumn{1}{|l|}{MAX}                         & \multicolumn{1}{l|}{7.3}                          \\ \midrule
\multicolumn{1}{|l|}{MEAN}                        & \multicolumn{1}{l|}{2.36}                         \\ \midrule
\multicolumn{1}{|l|}{STD DEV}                     & \multicolumn{1}{l|}{2.20}                         \\ \midrule
\multicolumn{1}{|l|}{\textbf{RMSE}}                & \multicolumn{1}{l|}{\textbf{1.78}}                         \\ \midrule
\end{tabular}
\caption{Distance Measurement Result}
\label{distance_table}
\end{table}

\section*{Conclusion and Future Works}
In this research, we proposed a landmark anchor-based distance measurement strategy to localize and guide moving nodes (say troops) in a battlefield environment where GPS is either unavailable or not desirable to use. We introduced two real-world datasets (MSTLandamrkv1 and MSTLandmarkStereov1) for landmark recognition and stereo distance measurement tasks respectively. We performed transfer learning on Yolov5 and fined-tuned the model to detect the landmark anchors in the images. Then we applied the Semi Global Stereo-matching algorithm that is integrated with the result of landmark detection to generate depth patches of detected landmark anchor. The depth patch is later aggregated to obtain the average distance to the landmark. 
We developed a method to store the distance between the mobile node and different landmarks within each mobile node as virtual coordinates, which can be used to perform node localization while traversing through a battlefield using trilateration in future works. We aim to augment our research to encompass the localization and tracking of moving nodes, thereby guaranteeing the safety and survival of troops traversing battlefield zones. By leveraging landmark-based virtual coordinates, we will establish a system that verifies whether troops remain within the safe path designated by the control center. Our efforts will focus on generating geometric constraints based on the proximity of landmarks and the virtual coordinates of individual nodes, enabling us to monitor troop movements and alert them of any deviations from the designated safe path.

\end{document}